\documentclass[letterpaper]{article}
\usepackage{icmlfiles/uai2020}
\usepackage[margin=1in]{geometry}
\usepackage{graphicx}
\usepackage{subfig}
\usepackage{booktabs} 
\usepackage[colorlinks=true,linkcolor = blue, urlcolor  = blue,citecolor = blue, anchorcolor = blue]{hyperref}
\usepackage{multicol}
\usepackage{enumitem}
\usepackage{tabularx}
\usepackage{array, caption, floatrow, makecell, booktabs}%
\setitemize{noitemsep,topsep=0pt,parsep=0pt,partopsep=0pt}
\usepackage{amsmath,amssymb}
\DeclareMathOperator{\E}{\mathbb{E}}
\usepackage{xcolor}
\usepackage[round]{natbib}
\setcitestyle{authoryear,round,citesep={;},aysep={,},yysep={;}}

\usepackage{times}

\newcommand{\an}[1]{{\color{black}{#1}}}
\title{Unsupervised Calibration under Covariate Shift}

\author{} % LEAVE BLANK FOR ORIGINAL SUBMISSION.
\author{ {\bf Anusri Pampari} \\
Computer Science Department \\
Stanford University\\
\And
{\bf Stefano Ermon}   \\
Computer Science Department \\
Stanford University    \\
}
\newtheorem{theorem}{Theorem}[section]

\newtheorem{lemma}[theorem]{Lemma}

\begin{document}

\maketitle

\begin{abstract}

A probabilistic model is said to be calibrated if its predicted probabilities match the corresponding empirical frequencies. Calibration is important for uncertainty quantification and decision making in safety-critical applications. While calibration of classifiers has been widely studied, we find that calibration is brittle and can be easily lost under minimal covariate shifts. Existing techniques, including domain adaptation ones, primarily focus on prediction accuracy and do not guarantee calibration neither in theory nor in practice. In this work, we formally introduce the problem of calibration under domain shift, and propose an importance sampling based approach to 
%adapt existing calibration methods to 
address it. We evaluate and discuss the efficacy of our method on both real-world datasets and synthetic datasets.
\end{abstract}
\section{INTRODUCTION}
\label{submission}

%\se{doesn't look in uai format. font is weird (and nicer than the standard uai one) }
%\an{now it looks okay? I am unable to make out the difference ouch, going blind lol. }
%{\color{blue} This para to introduce importance of confidence estimates etc}
%\se{better to use motivating examples like medical diagnosi where calibration is important. justice system, financial decisions, human robot interaction, etc.}
Machine learning models are increasingly being entrusted with complex decisions in many applications such as medical diagnosis \citep{triantafyllidis2019applications}, justice system \citep{berk2015machine}, financial decisions \citep{heaton2017deep}, human robot interaction \citep{modares2015optimized}, etc.  In all these applications, models must not only be accurate, but should also indicate confidence in their own predictions. Uncertainity quantification is  important for safety-critical applications and in decision making. This will better inform when the model's predictions are likely to be incorrect, and help in building trust with the user. For example, in medical diagnosis if the model is not confident about it's prediction, then the decision making should be passed on to a doctor. Additionally, humans have a natural cognitive intuition for probabilities \citep{cosmides1996humans}. \an{Calibrated probabilities provide an intuitive explanation to a model's predictions, making them interpretable.} %\se{better?} 

%Calibrated probabilities provide an additional bit of information about the model making them interpretable. \se{? is this meant in a technical sense as a bit of information?}
%\an{actually no}

%Add, \ste{uncertainty quantification important for safety, interpretability, used for decision-making}

%{\color{blue} This para to introduce the problem of calibration, what calibration means intuitively and concretely.}
Ideally, the confidence or probability associated with the predicted class label should reflect its ground truth occurrence likelihood. For example, suppose a diabetes risk prediction model predicts a chance of 70\% for a specific patient profile. Then, we expect that out of 100 similar patients, about 70 should have diabetes. Such a model is said to be \textit{calibrated}. Many existing machine learning models, such as SVMs, Gaussian processes, and Neural Networks, are not naturally calibrated \citep{guo2017calibration, bella2010calibration}, thus producing unreliable confidence estimates. This can, in turn, lead to bad decision making and reduce the trust in using these models.

\begin{figure}
  \centering
{\includegraphics[width=\linewidth]{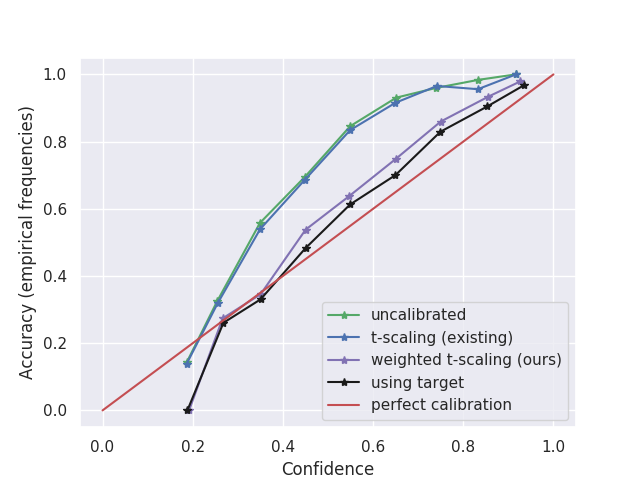}
}
  \caption{Reliability diagram for a LeNet-5 model trained using CDAN (SOTA domain adaptation technique) on MNIST  and tested on USPS as target data.}
  \label{main-fig}
\end{figure}

%{\color{blue} Limitations of current literature}
%\se{this might be misinterpreted. this is true if there is distribution shift after training}
Existing literature \citep{platt1999probabilistic, zadrozny2001obtaining, zadrozny2002transforming, bella2010calibration, guo2017calibration} introduces many post processing techniques to correct these miscalibrated models. However, they assume the availability of labeled held-out validation data drawn from the same distribution as the test data to achieve calibration. This assumption is violated in many real world scenarios in the following two ways. Firstly, the test dataset can have a different distribution due to covariate shift. This can happen, e.g., when the operating conditions at test time are slightly different. Secondly, labelled test data is often unavailable if distribution shift occurs after training. While several unsupervised domain adaptation methods \citep{chu2018survey, kouw2019review} propose solutions for correcting the accuracy of the models, there is no existing work to correct the effects of these circumstances on the confidence. For example, in Figure. \ref{main-fig}, we show how an existing calibration method temperature scaling (t-scaling) \citep{guo2017calibration}  can fail to calibrate a LeNet-5 model trained using CDAN (a SOTA domain adaptation model \citep{long2018conditional}) under domain shift. Here the model is trained on MNIST and has a prediction accuracy of 70\% on USPS dataset. 
% mention accuracy to say accuracy does not mean calibration
%\se{this might be misinterpreted. this is true if there is distribution shift after training}
%\ste{i thought you had found a 2019 neurips paper?} -\an{add the paper}

In this work, we introduce and investigate the problem of miscalibration under covariate shift. We demonstrate that existing models learnt using domain adaptation are poorly calibrated, showing that while current domain adaptation techniques account for accuracy, they do not consider calibration of the models. \an{We then propose a modification to the calibration optimization objective used by existing techniques. Our solution employs importance sampling to account for the difference in the training and testing distributions, thereby overcoming the inherent assumptions of the existing methods. }%We then propose a solution to overcome the assumptions made by existing calibration techniques and adapt them to work under covariate shift. \se{be more specific if you can, otherwise this might not be clear} %\ste{this might seem contradictory, since we just said there is no such method yet}Our solution employs importance sampling in a learned feature space to re-weight the training data.
Our proposed method can adapt any existing calibration method under covariate shift assumption without requiring any labeled data from the test distribution. In Figure. \ref{main-fig}, we show how our method (weighted t-scaling) adapts the use of t-scaling on source validation data to work under domain shift. We achieve close performance to perfect calibration or calibration obtained using labeled target data.

%We also show the reference lower bound \se{lower bound on what?} that can be obtained by \textit{using target} labels (if available) for calibration. \se{what's the takeaway?} 

To summarize our contributions,
\begin{itemize}
    \item  We introduce the problem of miscalibration under covariate shift, and show how existing domain adapted  models such as CDAN \citep{long2018conditional} remain uncalibrated in the target domain on using existing calibration methods;
    \item We propose an importance sampling based solution to address the problem of calibration under covariate shift. Our method requests no additional labels from the test distribution and can be used to adapt any calibration method;
    \item We use a discriminator trained on a domain-invariant feature layer of source and target to get density ratios for importance sampling.
\end{itemize}

\section{RELATED WORK}
\label{rwork}

%$Y \in {1,\ldots, K}$
\textbf{Background and Notation} Calibration can be described mathematically as follows. Suppose that we have some data, comprising of inputs $X \in R^d$ and labels $Y \in {1,\ldots, K}$, which follows the ground truth joint distribution $\pi(X, Y) = \pi(Y|X)\pi(X)$. Let $h(.)$ be a classifier learned for this data, i.e.  $ h: X \rightarrow [0,1]^K$ 
%\begin{equation*}
%   h(X) = \pi_{\theta}(Y|X),\ h: X %\rightarrow [0,1]^K.
%\end{equation*}
which for an input $x \sim X$, is the probability distribution over the $K$ classes in $Y$. The class with the maximum probability of occurrence is the prediction $\hat{Y}$, and its corresponding  probability is the confidence prediction $\hat{P}$. The classifier $h$ is said to be calibrated \citep{guo2017calibration} when,
\begin{equation}
    \mathbb{P}(\hat{Y}=Y | \hat{P}=p) = \text{p  }  \forall \text{ p }  \in [0,1]
\end{equation}
%\se{add a reference for this? i'm not even totally sure what's the right definition of calibration for multi-class classification, but can certainly imagine other variants}
%The classifier $h$ is said to be calibrated iff  $\pi_{\theta}(Y|X) =  \pi(Y|X)$, i.e. the predicted likelihood of a class  matches its empirical probability (citep). 
%\ste{this is not the usual def of calibration. this is a very strong property if required for all x. typically it's only required to hold "on average"}
%This error is inevitable when using finitely many samples to learn the classifier $h$. \se{aso because of model mismatch, optimization issues, etc}
Many existing classifiers do not naturally satisfy these requirements \citep{guo2017calibration, bella2010calibration}, and are therefore said to be miscalibrated. \an{This error is inevitable because of many reasons such as using finitely many samples to learn the classifier $h$, model mismatch from the true distribution, optimization issues, etc.} Existing calibration techniques \citep{platt1999probabilistic, zadrozny2001obtaining, zadrozny2002transforming, bella2010calibration, guo2017calibration}  reduce this error in post-processing steps to produce calibrated probabilities. A calibration model (parametrized by $\beta$) is applied over the uncalibrated classifier. Each method defines an approximate variant of the calibration error using a loss function of the form  $\E_{(x,y) \sim \pi(X,Y)}l(h(x),y;\beta)$ and learns the parameters $\beta$ as the minimizer of this loss. We discuss some of the popular calibration methods and their corresponding expected loss function briefly.

\textit{Platt Scaling:} \citep{platt1999probabilistic} is a parametric approach to calibration. The multi-class predictions of a classifier $h$ are used as features for a multionomial logistic regression model $f_{\beta}$, which is trained on the validation set to return probabilities.  The accuracy of the classifier $h$ can change when using the calibrated probabilities for prediction \citep{guo2017calibration}.

\textit{Temperature Scaling (t-scaling):} This method proposed by \citep{guo2017calibration} is popularly used for neural network calibration. It uses a single scalar parameter $\beta$ called the temperature for all classes.  Here, the calibrated probabilities do not affect the accuracy of the classifier $h$.
%$f_{\beta}$ here is defined to modify the predicted probabilities by raising its output entropy. \se{wont' be clear}

Parameters $\beta$ of the calibration model $f_{\beta}$ in both the above methods is optimized by using the NLL loss over the validation set. Hence,
\begin{equation}
    \E_{(x,y) \sim \pi(X,Y)}l(h(x),y,;\beta) = \sum_{i=1}^{n} \sum_{j=1}^{K} y_i^{j} log(f_{\beta}(h(x_i)^{j})
    \label{plat}
\end{equation}
where n is the number of samples drawn from the joint distribution $(h(X),Y)$ and $Y$ is represented as one hot vector of size $K$.

\textbf{Quantifying miscalibration:}
The common  metrics used to report calibration performance are Expected Calibration Error or ECE \citep{guo2017calibration} and reliability diagrams \citep{degroot1983comparison, niculescu2005predicting}. We briefly describe both these measures and use it in our work to report calibration performance.

We start with grouping confidence predictions  $\hat{p}_i$ into $M$ interval bins (each of size $1/M$). Let $B_m$ be the set of indices of samples whose prediction confidence falls into the interval $I_m = (\frac{m-1}{M},\frac{m}{M}].$ We define accuracy of bin $B_m$ as,
\begin{equation*}
    acc(B_m) = \frac{1}{|B_m|} \sum_{i \in B_m} \textbf{1}(\hat{y}_i = y_i)
\end{equation*}
where $\hat{y}_i$ and $y_i$ are the predicted and true class labels for
sample i. We also define the average confidence within bin $B_m$ as,
\begin{equation*}
    conf(B_m) = \frac{1}{|B_m|} \sum_{i \in B_m} \hat{p}_i
\end{equation*}
where $\hat{p}_i$ is the confidence for sample i.
%Mention the number of bins used in the experimentation

\textit{Expected Calibration Error (ECE):} \citep{guo2017calibration} define ECE as a weighted average of the  bins’ accuracy/confidence difference.
\begin{equation*}
    ECE =  \sum_{m = 1}^{M} \frac{|B_m|}{n} |acc(B_m) - conf(B_m)|
\end{equation*}
where n is the number of samples. Lower ECE indicates better calibration. %\an{Write a few more lines how to interpret this results}

\textit{Reliability diagrams} are visual representation of model calibration \citep{degroot1983comparison, niculescu2005predicting} as shown in Figure. \ref{main-fig}. These diagrams plot accuracy or the empirical frequency $acc(B_m)$  as a function of confidence $conf(B_m)$ for each bin $B_m$. So the x-axis here ranges from $[0,1]$ and is divided into M intervals. If the model is perfectly calibrated the diagram should plot the identity function. Any deviation from a perfect diagonal represents miscalibration.

%Existing calibration methods, define ECE and reliability diagram and all the existing methods.
%None considered the domain shift setting.
%https://arxiv.org/abs/1809.08159

\textbf{Limitations of existing work:} Existing calibration methods as discussed above rely on the evaluation of the loss function $\E_{(x,y) \sim \pi(X,Y)}l(h(x),y,;\beta)$, which necessitates the need of labeled held-out validation data. Here the methods inherently assume that the train, validation and test data is drawn from the same distribution $\pi(X,Y)$. This is violated in many real world scenarios where covariate shift occurs (discussed in Section. \ref{prob}). \citep{snoek2019can} empirically show that deep NN are uncalibrated under domain shift.

%\se{it might make sense to have a brief background section on covariate shift, and domain adaptation techniques. expanding the paragraph below and putting before the limitations of existing wok}
The effect of covariate shift on classifiers predictive performance and various solutions to address it has been studied under unsupervised domain adaptation literature \citep{chu2018survey, kouw2019review}. They assume availability of labeled train data and assume no labels on the test data. The performance of these models is measured using accuracy, while not considering the calibration of the models.  Our work introduces this issue for domain adaptation models.  We show how existing methods fail to calibrate domain adaptation models and  provide a simple modification to  adapt any existing calibration technique to dataset shift.

\section{MISCALIBRATION UNDER COVARIATE SHIFT}
\label{prob}

\begin{figure*}
  \centering
  \mbox{
    \subfloat[\label{subfigure label1}]{\includegraphics[width=0.32\linewidth]{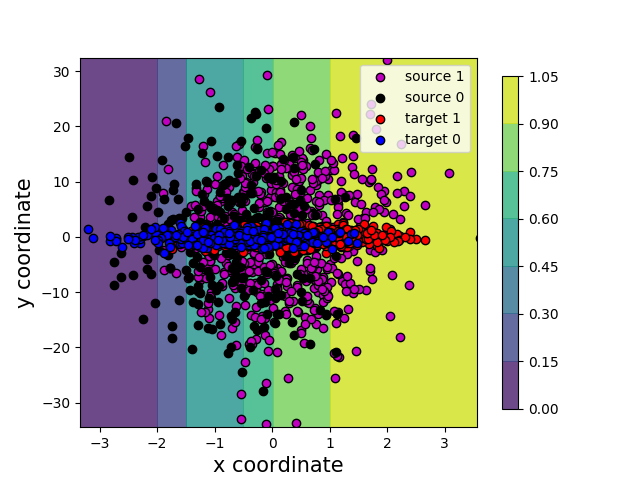}}\quad
    \subfloat[\label{subfigure label2}]{\includegraphics[width=0.32\linewidth]{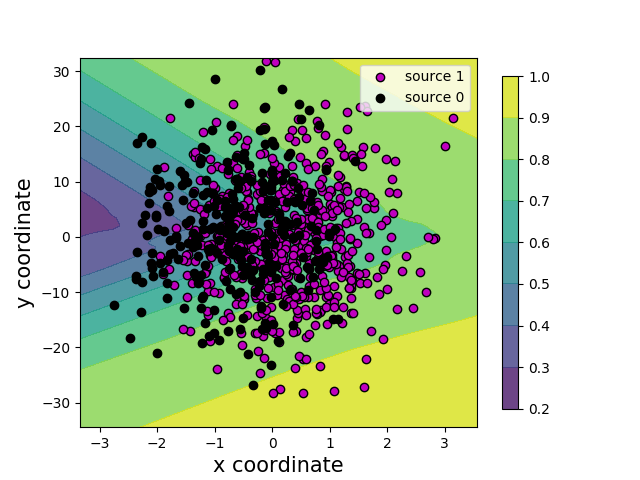}}\quad
    \subfloat[\label{subfigure label3}]{\includegraphics[width=0.32\linewidth]{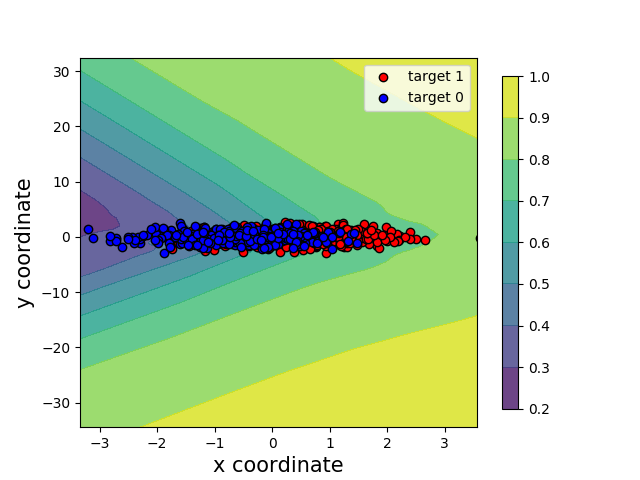}}
  } 
    \mbox{
    \subfloat[\label{subfigure label4}]{\includegraphics[width=0.32\linewidth]{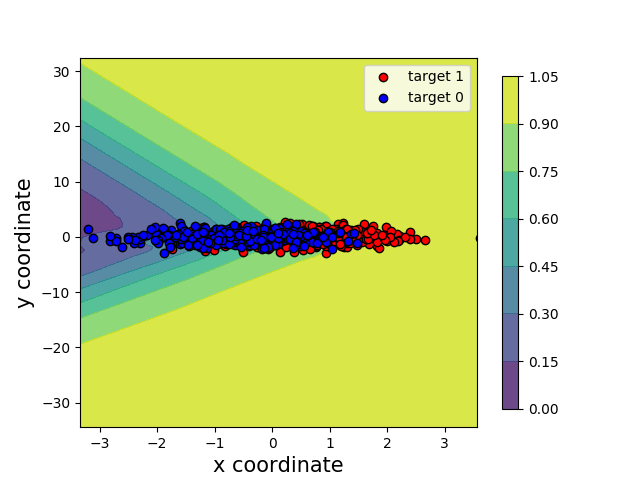}}\quad
    \subfloat[\label{subfigure label5}]{\includegraphics[width=0.32\linewidth]{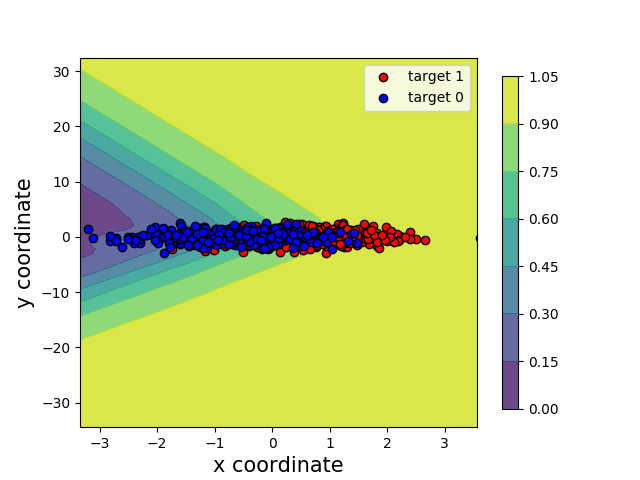}}\quad
    \subfloat[\label{subfigure label6}]{\includegraphics[width=0.32\linewidth]{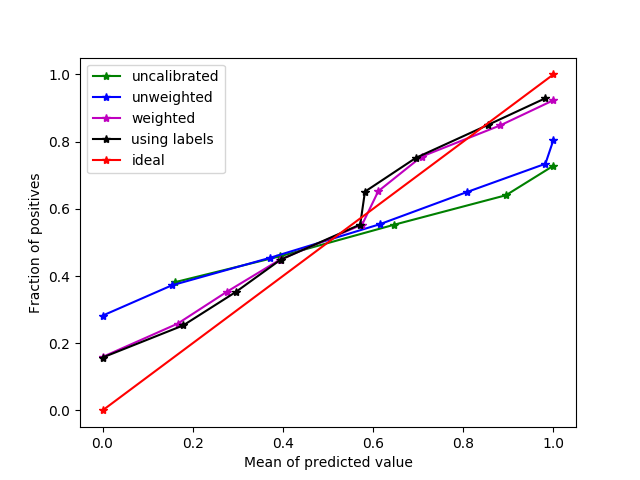}}
  }
  \caption{Comparing true probability distribution $\pi(Y=1|X)$ with $\pi_\theta(Y=1|X)$ (output of classifier $h$) obtained after post-hoc calibration.  (a) true $\pi(X)$ and $\pi(Y|X)$ (b) $\pi(X)$ of source and $\pi_\theta(Y|X)$  after calibration using source  (c) $\pi(X)$ of target and $\pi_\theta(Y|X)$  after calibration using source  (d) $\pi(X)$ of source and $\pi_\theta(Y|X)$  after calibration using target (e) $\pi(X)$ of source and $\pi_\theta(Y|x)$  after calibration using our weighted method. (f) Reliability diagram on target}
  \label{prob-cont}
  \vspace{-0.15in}
\end{figure*}

In this section, we discuss covariate shift and how it affects calibration using a synthetic example shown in Figure. \ref{prob-cont}. We organize this section as follows- (1) we formalize the assumption of covariate shift (2) we consider a case of a miscalibrated classifier (3)  We attempt to calibrate the classifier using existing techniques on source validation data and (4) We assume the availability of target labels and show how the calibration performance can differ from doing calibration using source as in (3). 

\textbf{Covariate shift assumption:} Consider that $\pi^{tr}(X,Y)$ represents the joint distribution of inputs and outputs of the training data and $\pi^{te}(Y|X)$ represents the same for the testing data.  Under covariate shift, we assume that $$\pi^{tr}(X) \neq \pi^{te}(X)$$ and $$\pi^{tr}(Y|X) = \pi^{te}(Y|X),$$ i.e the input distribution changes between train and test data (covariate denotes input), while the conditional distribution of the outputs given the inputs $\pi(Y|X)$ remains unchanged. This is illustrated in Figure. \ref{prob-cont}(a) using two multivariate Gaussian distributions with different co-variance matrices as our initial distribution for $\pi^{tr}(X)$ for source and $\pi^{te}(X)$ for target.  We consider a binary classification task where $\pi(Y=1|X)$ is the same for both source and target and changes as a function of the x-coordinate. This results in the difference of joint distribution between train and test, $\pi^{tr}(X,Y) \neq \pi^{te}(X,Y)$. The resulting labeled source and target data are also highlighted as source 1, source 0 and target 1, target 0.

\textbf{Classifier mis-calibration:}  To emulate a realistic setting, we mis-specify the classifier $h$ by training a non-linear MLP classifier on finite samples from the source data. This allows us to have a situation for which  probability distribution learnt by $h$ significantly deviates from the true $\pi(Y|X)$. \an{If $h$ was capable of learning the true relationship, the model would be calibrated both on source and target.}

%\textbf{Correcting the calibration with existing techniques:} We then attempt to calibrate $h$ using an existing calibration technique isotonic regression \an{(cite)}, resulting in the calibrated probability distribution shown in Fig. \ref{prob-cont}(b) for source and in \ref{prob-cont}(c) for target. The loss function used in the current calibration methods (as discussed in Sec.\ref{rwork}) is defined on a held out source validation data  i.e $\E_{(x,y) \sim \pi^{tr}_{(X,Y)}}l(h(x),y,;\beta)$ assumed to have the same distribution as the test data.  These methods are ideal when the train and test distributions are identical. However, if $\pi^{tr}(X,Y) \neq \pi^{te}(X,Y)$, it follows that the expected loss function to be minimized  is different for train and test distribution, i.e. $\E_{(x,y) \sim \pi^{tr}_{(X,Y)}} [l(h(x),y,;\beta)]$ $ \neq \E_{(x,y) \sim \pi^{te}_{(X,Y)}} [l(h(x),y,;\beta)]$. On using calibration methods derived using the train data over the shifted test data, the confidence estimates given by the model are no longer reliable. 

\textbf{Correcting the calibration with existing techniques:} We then attempt to calibrate $h$ using an existing calibration technique called isotonic regression \citep{zadrozny2002transforming} resulting in the calibrated probability distribution shown in Figure. \ref{prob-cont}(b) for source and in \ref{prob-cont}(c) for target. \an{Here we notice that $h$ is calibrated on source, but not on target. For example consider the 0.7-0.8 probability band. Here, 70-80\% points are positive in source showing calibration whereas nearly 100\% of points in the target are positive (red), showing miscalibration.}. The loss function used in the current calibration methods (as discussed in Section. \ref{rwork}) is defined on a held out source validation data assumed to have the same distribution as the train data, i.e $\E_{(x,y) \sim \pi^{tr}_{(X,Y)}}l(h(x),y,;\beta)$ is the evaluated loss function. These methods are ideal when the train and test distributions are identical. However, if $\pi^{tr}(X,Y) \neq \pi^{te}(X,Y)$, it follows that the expected loss function to be minimized is different for train and test distribution, i.e. $\E_{(x,y) \sim \pi^{tr}_{(X,Y)}} [l(h(x),y,;\beta)]$ $ \neq \E_{(x,y) \sim \pi^{te}_{(X,Y)}} [l(h(x),y,;\beta)]$. On using calibration methods derived using the train data over the shifted test data, the confidence estimates given by the model are no longer reliable. 
%\se{maybe emphasize that it is calibrated on source, but not on target. for example, nearly 100 percent of points in the 0.7-0.8 band are positive (red)} 

\textbf{Calibration using target data:} One can solve this by obtaining labeled data under the test distribution $\pi^{te}(X,Y)$ and directly computing the expected loss function over the test data, $\E_{(x,y) \sim \pi^{te}_{(X,Y)}} [l(h(x),y,;\beta)$. However, we often do not not have access to the label information on the test data.  Here for illustration, we assume the availability of labeled test data for calibration  and show the resulting probability distribution in Figure. \ref{prob-cont}(d) for the target data. The difference in the resulting probability distribution in Figure. \ref{prob-cont}(c) and Figure. \ref{prob-cont}(d), clearly show how the calibrated probability distributions differ  when using source or target data for calibration. Further, Figure. \ref{prob-cont}(f) shows quantitatively using reliability diagram that using source data for calibration on target can perform worse than an uncalibrated model moving it further away from perfect calibration.

%\se{this seems to be logically separate. i would put in a separate subsection, and expand a bit}

\textbf{Mis-calibration in domain adapted classifier:} The above discussion also extends to the case when $h$ is learnt using existing domain adaptation techniques.  In addition to labeled source data, these techniques also use the unlabeled target data to learn the classifier $h$. This reduces overfitting of the learnt classifier on the source labeled data, and hence improves the generalization (or predictive accuracy) on the unseen  target data. However, these models can still remain uncalibrated in the target domain. This is inevitable because we learn $h$ from finite source data, or simply due to optimization issues.  In Figure. \ref{main-fig} we show a reliability diagram  of domain adapted classifier (CDAN on LeNet-5) on USPS dataset, the classifier is trained on labeled MNIST and achieves an accuracy of 70\% on USPS. We notice that uncalibrated classifier is far from perfect. Even after using existing calibrations methods like t-scaling on source validation data, we notice that the model still remains uncalibrated.

Both the synthetic and  domain adapted examples discussed above, show the performance gap between perfect calibration (or reference calibration obtained using labeled target) and calibration obtained by using source data. We seek to close this performance gap without requesting new labeled data from the target.
\section{IMPORTANCE SAMPLING FOR CALIBRATION UNDER COVARIATE SHIFT}

To address the problem of miscalibration under covariate shift discussed in Section. \ref{prob}, we introduce an importance sampling approach for estimating the calibration loss. For this, we assume access to labeled training data $(x,y) \sim \pi^{tr}(X,Y)$, and unlabeled test data $x \sim \pi^{te}(X)$. A classifier $h$ is assumed to be trained either using only the labeled train data, or by using existing unsupervised domain adaptation techniques.  Our objective is to ensure that the classifier $h$ is calibrated on the test distribution. We describe our approach and the intuition behind it.  

 Consider the calibration loss defined over the test distribution $\E_{(x,y)\sim \pi^{te}(x,y)}l(h(x),y,;\beta)$. This cannot be computed using samples drawn from $\pi^{te}(X,Y)$ since we do not have access to the labels $Y$ from the test distribution. However, note that we have access to samples drawn from $\pi^{tr}(X,Y)$ and hence the calibration loss over the training distribution can be computed as $\E_{(x,y)\sim \pi^{tr}(X,Y)}l(h(x),y,;\beta)$. Hence, we seek to adapt the calibration loss defined on the training distribution to formulate the calibration error on the test distribution. This can be done using importance sampling in the following way:-
\begin{eqnarray*}
    \lefteqn{\E_{(x,y)\sim \pi^{te}(X,Y)}l(h(x),y,;\beta) } \\
    &   = \int_{x} \int_{y} l(h(x),y;\beta) \pi^{te}(x,y) dx dy \\
    &  =  \int_{x} \int_{y} l(h(x),y;\beta) \frac{\pi^{te}(x,y)}{\pi^{tr}(x,y)} \pi^{tr}(x,y) dx dy \\
    &  =  \int_{x} \int_{y} l(h(x),y;\beta) \frac{\pi^{te}(x)\pi^{te}(y|x)}{\pi^{tr}(x)\pi^{tr}(y|x)} \pi^{tr}(x,y) dxdy
\end{eqnarray*}

%\se{tidy up the equations, formatting} 

Using the covariate shift assumption, we have $\pi^{tr}(X) \neq \pi^{te}(X)$ and $\pi^{tr}(Y|X) = \pi^{te}(Y|X)$. From these assumptions, it follows that:
\begin{eqnarray*}
    \lefteqn{\E_{(x,y)\sim \pi^{te}(X,Y)}[l(h(x),y,;\beta)]} \\
    &   = \int_{x} \int_{y} l(h(x),y,;\beta) \frac{\pi^{te}(x)}{\pi^{tr}(x)} \pi^{tr}(x,y) dx dy \\
    &   = \E_{(x,y)\sim \pi^{tr}(X,Y)} \frac{\pi^{te}(x)}{\pi^{tr}(x)} l(h(x),y,;\beta)
\end{eqnarray*}

The above result is summarized in Theorem \ref{tgt-calib},

\begin{theorem}
\label{tgt-calib}
The calibration loss with covariate shift correction on the test data is equivalent to the density ratio weighted calibration loss on the training data, i.e
\begin{eqnarray*}
    \lefteqn{\E_{(x,y)\sim \pi^{te}(X,Y)}[l(h(x),y,;\beta)] } \\
    &   = \E_{(x,y)\sim \pi^{tr}(X,Y)}[\gamma(x)l(h(x),y,;\beta)],
\end{eqnarray*} where $\gamma(x) = \frac{\pi_{te}(x)}{\pi_{tr}(x)}$ is the density ratio and $supp(\pi^{te})$ $\supset$ $supp(\pi^{tr})$ where $supp(\pi^{tr})$ = $\{x|\pi^{tr}(x) = 0\}$
\end{theorem}

Weighting the train data with density ratios given by $\gamma(x)$ is an importance sampling approach.  By increasing the relative weight of those regions of the training distribution which also have a high density under the test distribution, we adapt $\pi^{tr}$ to represent $\pi^{te}$. We can observe the qualitative behaviour of calibration when using our method in the following way. Consider the synthetic data example and the isotonic regression calibrator trained on the source data in Section. \ref{prob}. We incorporate the weighted calibration loss for isotonic regression to optimize the  calibrator on the source data. Here, we use the ground truth density ratios computed from the known distributions. The resulting probability distribution obtained in Figure. \ref{prob-cont}(e) using our method is similar to the probability distribution obtained by using target labels in Figure.  \ref{prob-cont}(d). We also note from the reliability diagram in Figure.  \ref{prob-cont}(e) that the performance after using weighted calibration loss is closer to perfect-calibration.

%\se{we should be explicit that the pi are unknown, but we have samples. mention some of the techniques for estimating importance weights in a likelihood-free way}

\an{In order to estimate density ratios  $\gamma(x)$, we require knowledge of the true data distribution $\pi(x)$ for train and test data, which is unknown. However, we typically  have sampling access
to $\pi(x)$ via finite datasets which we use to estimate the density ratios in a likelihood-free (LF) way. Examples of some LF estimators include  nearest neighbour \citep{kremer2015nearest}, discriminative estimation \citep{bickel2007discriminative} etc.} 
% KLIEP \citep{sasaki2015direct}
We can, in principle, estimate this ratio directly in the original input space. However, when the inputs are high dimensional, the estimated loss  may suffer from large estimation variances because of greater divergence between the distributions $\pi^{te}$ and $\pi^{tr}$ \citep{snoek2019can}. We further elaborate on this observation and discuss solutions to address it in the subsequent subsections.

%\se{maybe move before the discussion on how to estimate importance weights,}

\subsection{FEATURE REPRESENTATION FOR IMPORTANCE SAMPLING}

In this section, we discuss practical difficulties in applying importance weighted calibration and introduce a method to address some of these difficulties by using a suitable feature representation. There are two primary concerns in using importance weighted calibration on real train $\pi^{tr}$ and test $\pi^{te}$ distributions.
\begin{itemize}
    \item[1] \textbf{Accuracy of estimation}: The variance of calibration loss estimate in  Theorem \ref{tgt-calib}, and hence the accuracy of the calibration is affected by the divergence between $\pi^{te}$ and $\pi^{tr}$. This relation is discussed  in \citep{cortes2010learning} and summarized in Lemma \ref{variance} as follows:  
\begin{lemma} The variance of importance weighted calibration loss is bounded by the Renyi divergence $d_{\alpha}$,
\begin{eqnarray*}
    \lefteqn{Var_{\pi^{tr}}[l_\gamma] = \E_{\pi^{tr}}[(l_\gamma)^2]) - (\E_{\pi^{tr}}[l_\gamma])^2} \nonumber \\
   &   \leq d_{\alpha+1} (\pi^{te}||\pi^{tr}) (\E_{\pi^{te}}[l_\gamma)])^{1-\frac{1}{\alpha}} - (\E_{\pi^{te}}[l_\gamma))^2
\end{eqnarray*}
where Renyi divergence $d_{\alpha+1} (\pi^{te}||\pi^{tr})$ = $[\sum_x \frac{\pi^{te}(x)^{\alpha+1}}{\pi^{tr}(x)^{\alpha}} ]^{\frac{1}{\alpha}} $ where hyperparameter $\alpha > 0$.
\label{variance}
\end{lemma} 

From Lemma. \ref{variance} it is clear that the smaller the divergence, the better the chance of getting an accurate estimate of the calibration loss, in turn affecting the accuracy of the final calibrator. \\

\item[2] \textbf{Unbounded/ undefined density ratios}: The support of the train distribution might not contain the support of the test distribution as required by  Theorem \ref{tgt-calib}. When this is violated, the density ratios can grow to infinity thus resulting in a undefined or incorrect estimate. 

\end{itemize}

We address both these concerns using a method similar to \citep{you2019towards}, by estimating importance weights using domain-invariant features instead of the original covariates. Let $\pi^{tr}_f$ and $\pi^{te}_f$ be the domain-invariant feature distributions of the train and the test data respectively, we step from the input space to the feature space and estimate $\gamma_f(x) = \frac{\pi^{te}_f(x)}{\pi^{tr}_f(x)}$ instead of $\gamma(x) = \frac{\pi_{te}(x)}{\pi_{tr}(x)}$. This ensures that the variance of  calibration loss estimate is bounded, because by using domain-invariant features we have $ d_{\alpha+1} (\pi^{te}_f||\pi^{tr}_f)$ smaller than $ d_{\alpha+1} (\pi^{te}||\pi^{tr})$.  Furthermore, the assumption on the support of $\pi^{tr}$ in $\pi^{te}$ can hold well in the learned feature space because of increased overlap in the distributions compared to the covariate space. 

Note that by estimating the importance weights in the domain-invariant feature space we can only reduce the distribution divergence, and never completely eliminate it to zero. Hence we can only expect to improve over the bias created by original unweighted calibrator. Perfect calibration close to using the target labels is not guaranteed.
\subsection{DENSITY RATIO  ESTIMATION}
\label{ddre}

To compute density ratios we adopt an approach similar in \citep{bickel2007discriminative, you2019towards} where a discriminator is used to distinguish or classify  source  samples (with label d=1)  from  target samples (with label d=0). Under this model $\gamma_f(x) = \frac{\pi^{te}_f(x)}{\pi^{tr}_f(x)}  = \frac{P (d=1)}{P(d=0)} \frac{P (d=0|x)}{P(d=1|x)}$  where density ratio estimation is decomposed into two parts - (1) $\frac{P (d=0|x)}{P(d=1|x)}$ which can be estimated by a discriminative model to distinguish source and target samples.  The model here is trained on the domain-invariant feature representation of source and target.  and (2)  $\frac{P (d=1)}{P(d=0)}$ - is a constant value that can be estimated with the sample sizes of both domains.

\textbf{Practical considerations:}  The importance weights learnt by the discriminator may differ from the true density ratios. This can happen because, (1) the divergence between train and test is not completely zero leading to high variance (Lemma. \ref{variance}) and (2) training on finite samples from source and target data leads to  over-fitting  of the discriminator on some features. This results in highly confident predictions and hence small importance weights. We follow \citep{grover2019bias} to offset these challenges using the following techniques - self-normalization, flattening, and clipping.

 \section{EXPERIMENTAL SETUP}

% The model being calibrated is LeNet-5 trained on subclasses of CIFAR-10 present in different mixing ratios in source and target. 
% Brackets show the standard deviation across 10 different train-test splits.

\begin{figure*}
  \centering
  \mbox{
    \subfloat[\label{subfloat label1}]{\includegraphics[width=0.32\linewidth]{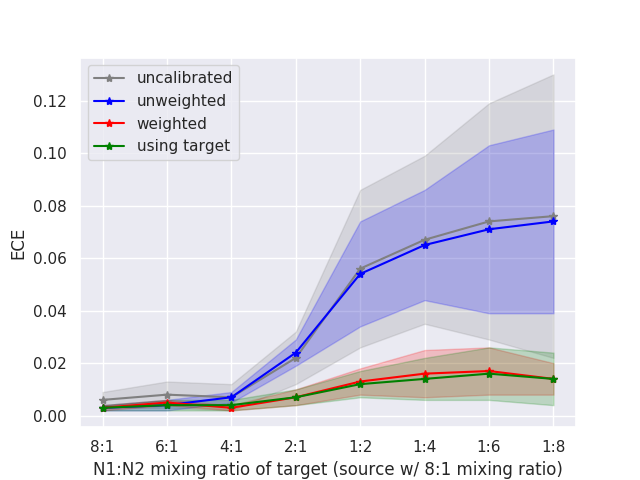}}\quad
    \subfloat[\label{subfloat label2}]{\includegraphics[width=0.32\linewidth]{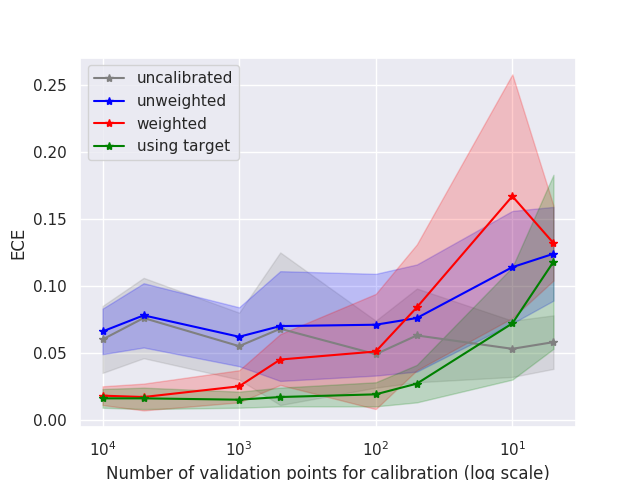}}\quad
    \subfloat[\label{subfloat label3}]{\includegraphics[width=0.32\linewidth]{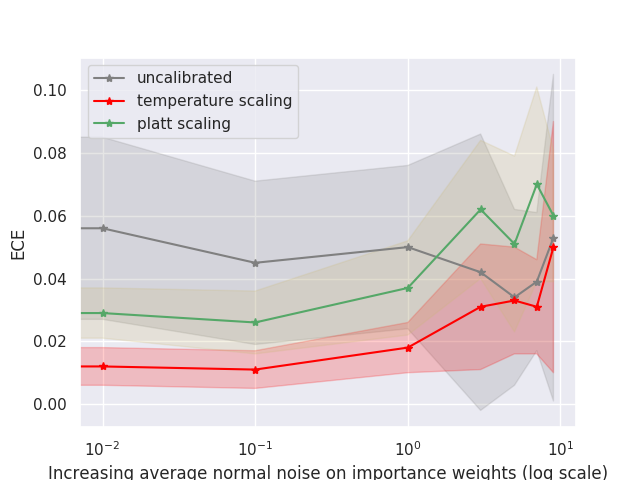}}
  }
  \caption{Figure showing different parameters that effect calibration. (a) Increasing divergence between source and target (b) Increasing number of samples used for calibration (c) Increasing noise in the ground truth importance weights}
  \label{fig:syn}
\end{figure*}

%The source models are already calibrated using these techniques and hence we do not report them.
In this section, we evaluate the efficacy of our proposed importance weighting technique in adapting two post-hoc calibration methods, Platt scaling and temperature scaling (t-scaling), to handle domain shifts. We use the Expected Calibration Error (ECE) discussed in Sec. \ref{rwork} to measure the calibration performance on the target data.

%We consider experiments on both pseudo-real and real world datasets.

We compare the performance of our calibration method (Weighted) to three baselines:\\ 
(1) \textbf{Uncalibrated}, i.e. the source classifier as is without any post-hoc calibration;\\
(2) \textbf{Unweighted}, i.e. the post-hoc calibrator is trained on the source domain;\\
(3) \textbf{Using target or target-calibrated} i.e. the post-hoc calibrator is trained on the labeled target domain. This can be considered as a gold standard (requiring labels from target domain), i.e. a   lower-bound on the calibration error for the target data. 
 \begin{table}[!ht]
\setlength\tabcolsep{2.5pt}
\begin{tabularx}{\textwidth}{@{} l c c c @{}}
\toprule
Dataset & CIFAR-10 classes & source ratio & target ratio \\
\midrule
S1 $\rightarrow$ T1 & 2\&7 & 1:4 & 4:1 \\
S2 $\rightarrow$ T2 & 1\&8 & 2:5 & 3:4\\
S3 $\rightarrow$ T3 & 3\&4  & 5:1 & 1:3\\
S4 $\rightarrow$ T4 & 6\&9  & 2:3 & 5:1\\
\bottomrule
\end{tabularx}
\setlength\tabcolsep{2.5pt}
\begin{tabularx}{\textwidth}{@{} l l l l l  @{}}
\toprule
Method & S1 $\rightarrow$ T1 & S2 $\rightarrow$ T2 & S3 $\rightarrow$ T3 & S4 $\rightarrow$ T4  \\ 
\midrule
Uncalibrated & 0.134  &  0.019 &  0.142 & 0.020 \\ \hline
Unweighted & 0.163 & 0.018 & 0.204 & 0.026\\
Weighted & \textbf{0.040} & 0.023 & \textbf{0.042} &   \textbf{0.017} \\
Using target &  0.037 & 0.024 & 0.041 &  0.016\\ \hline \hline
Unweighted & 0.124 & 0.007 &  0.144 & 0.013 \\
Weighted &  \textbf{0.030} & 0.005 & \textbf{0.030} & \textbf{0.007}   \\
Using target &  0.027  & 0.005 &  0.029  & 0.007 \\
\bottomrule
\end{tabularx}
\caption{ECE scores of Platt (top) and t-scaling (bottom), comparing baselines and the proposed method on pseudo-synthetic datasets. The weighted method uses ground truth density ratios for calibration.}
\label{tab:syn}
\end{table}
 
The rest of the section is organized as follows. First, we study the behaviour of the calibration on pseudo-real datasets. This paradigm allows us to control the density ratios and analyze how the performance is affected by it. Then, we apply this technique on real world datasets. Here, we derive the importance weights using the discriminative density ratio estimation method (Sec. \ref{ddre}) and use them for weighted calibration. The classifiers used here include both ImageNet pre-trained ResNet50 \citep{he2016deep} models trained only on the labeled source data and popular domain adapted models such as CDAN \citep{long2018conditional} which use both labeled source data and unlabeled target data.

\begin{table*}
\subfloat[0.55\linewidth][Office-31 dataset trained using pre-trained Resnet50.]{
    \setlength\tabcolsep{4pt}
    \begin{tabularx}{0.55\linewidth}{@{} l l l l l l l @{}}
    \toprule
    Method&A$\rightarrow$D&A$\rightarrow$W&D$\rightarrow$A&D$\rightarrow$W&W$\rightarrow$A&W$\rightarrow$D\\
    \midrule
    Uncalibrated & 0.038 & 0.036 & 0.147 & 0.158 & 0.045 & 0.096 \\ \hline
    Unweighted   & 0.082 & 0.06  & 0.278 & 0.093 & 0.199 & 0.085 \\
    Weighted & 0.125 & \textit{0.051} & \textbf{0.06} & \textbf{0.041} & \textit{0.052} & \textbf{0.083} \\
    Using target & 0.105 & 0.062 & 0.057 & 0.06  & 0.026 & 0.013 \\ \hline \hline
    Unweighted   & 0.040 & 0.045 & 0.145          & 0.028 & 0.136                   & 0.023         \\
    Weighted     & 0.047 & 0.081 & \textbf{0.134} & 0.031 & \textit{\textbf{0.039}} & \textbf{0.02} \\
    Using target  & 0.047 & 0.030 & 0.031          & 0.029 & 0.028                   & 0.013        \\
    \bottomrule
    \end{tabularx}
    \label{tab:office_31}
}
\hfill
\subfloat[0.5\linewidth][Digits dataset, domain adapted using CDAN.]{
\setlength\tabcolsep{4pt}
    \begin{tabularx}{0.4\linewidth}{@{} l l l @{}}
    \toprule
    Method&U$\rightarrow$M&M$\rightarrow$U\\
    \midrule
    uncalibrated & 0.309          & 0.206          \\ \hline
    unweighted   & 0.134          & 0.258          \\
    weightes     & \textbf{0.132} & \textbf{0.077} \\
    using target & 0.035          & 0.023    \\ \hline \hline
    unweighted   & 0.154          & 0.196         \\
    weighted     & 0.108 & \textbf{0.13} \\
    using target & 0.140           & 0.042 \\
    \bottomrule
    \end{tabularx}
    \label{tab:digits}
}
%\vspace{-0.05in}
\caption{ECE scores of Platt scaling (top) and t-scaling (bottom), comparing  the baselines and the proposed method. Importance weight are estimated using discriminator.}
\label{bigtable}
\end{table*}

\begin{table*}
\setlength\tabcolsep{2.5pt}
\begin{tabularx}{\textwidth}{@{} l l l l l l l l l l l l l  @{}}
\toprule
Method&Ar$\rightarrow$Cl&Ar$\rightarrow$Pr&Ar$\rightarrow$Rw&Cl$\rightarrow$Ar&Cl$\rightarrow$Pr&Cl$\rightarrow$Rw&Pr$\rightarrow$Ar&Pr$\rightarrow$Cl&Pr$\rightarrow$Rw&Rw$\rightarrow$Ar&Rw$\rightarrow$Cl&Rw$\rightarrow$Pr\\
\midrule
Uncalibrated & 0.114 & 0.091 & 0.119 & 0.135 & 0.133 & 0.113 & 0.098 & 0.158 & 0.021 & 0.124 & 0.102 & 0.027 \\ \hline
Unweighted & 0.109 & 0.098 & 0.093 & 0.104 & 0.077 & 0.099 & 0.306 & 0.197 & 0.133 & 0.083 & 0.159 & 0.148 \\
Weighted &\textbf{0.093} &
  \textbf{0.081} &
  0.093 &
  \textbf{0.071} &
  \textbf{0.051} &
  \textbf{0.068} &
  \textbf{0.085} &
  \textit{0.173} &
  \textit{0.081} &
  \textbf{0.067} &
  \textit{0.132} &
  \textit{0.059} \\
Using target & 0.092 & 0.060  & 0.029 & 0.052 & 0.076  & 0.053 & 0.028 & 0.134 & 0.069 & 0.035 & 0.074 & 0.05 \\ \hline \hline
Unweighted & 0.145 & 0.033 & 0.038 & 0.041 & 0.079 & 0.078 & 0.038 & 0.287 & 0.104 & 0.117 & 0.167 & 0.027 \\
Weighted & \textit{0.128} & \textbf{0.015} & 0.064 & 0.069 & \textbf{0.064} & \textbf{0.057} & 0.085 & \textit{0.268} & 0.108 & \textbf{0.104} & 0.173 & 0.038 \\
Using target & 0.046 & 0.034 & 0.043 & 0.048 & 0.026 & 0.040  & 0.049 & 0.038 & 0.029 & 0.019 & 0.036 & 0.026 \\
\bottomrule
\end{tabularx}
%\vspace{-0.1in}
\caption{Office-home, domain adapted using using CDAN. ECE scores of Platt scaling (top) and t-scaling (bottom), comparing baselines and the proposed method. Importance weight are estimated using discriminator. }
\label{tab:office_home}
\end{table*}

 \subsection{PSEUDO-REAL WORLD EXPERIMENTS}
\label{prdataset}

We construct synthetic datasets using the CIFAR-10 dataset which consists of 60K color images distributed equally across ten object classes \citep{krizhevsky2009learning}. We randomly pick two classes and collect samples from these two to define a binary classification task. We vary the mixing ratio of the two classes, thereby creating datasets with covariate shift. Sample values used in experiments for source  and target domain are documented in Table~\ref{tab:syn} as ($S_i \rightarrow T_i$). For example, if the source consists of $1:4$ ratio of class 1 and class 2, then the target with ratio of $4:1$ for the classes can be seen to have a covariate shift. To create ratio values greater than one, we duplicate the data points of the class. Our method of construction automatically gives us the ground truth importance weights from the mixing ratios of source and target data. In the previous example, the source points from class 1 have an importance weight of 4 and the source points in class 2 have an importance weight of $(1/4) =$0.25 for the given target. %\se{might want to explain this better and worked it out}

Using these ground truth importance weights we perform a weighted calibration of a LeNet-5 classifier trained on the labeled source data.  We use 70\% of the source data for training the classifier and 30\%  as validation data for calibration. For testing, we use 70\% of the target data to compute the ECE and 30\% as validation data for target-calibrated model. We consider 10 different train, validation, and test splits and report the mean in Table. \ref{tab:syn} (standard deviation in supplementary). We note that the weighted calibration significantly outperforms unweighted and uncalibrated models except in $S_2 \rightarrow T_2$ where the source ratio is close to the target ratio. We further make use of this setting to empirically study the effect of the following parameters on the calibration performance (ECE) in the target domain.

 \textbf{Domain shift between source and target:} We consider two classes of CIFAR-10 and fix the source class ratio to $8:1$ and the calibration method to t-scaling. We then change the target ratios as $N1:N2$. Here as the value of $N1$ decreases and $N2$ increases the domain shift of target compared to the source increases.  In Figure. \ref{fig:syn}(a), we see that performance gap in ECE between unweighted source calibration and using target for calibration increases as the datashift increases. The weighted calibration performs as well as using target labels for calibration (lower bound), even though our method  doesn't have access to any target labels.

\textbf{Number of validation samples used for training the calibrator:}  We vary the number of validation points used for t-scaling in Figure. \ref{fig:syn}(b) and keep the remaining parameters fixed.  We note that the weighted ECE performance may worsen compared  to unweighted calibration at significantly smaller validation sample size. Also, the performance of target calibration and unweighted source calibration itself may degrade with decreasing validation samples. The threshold of sample size for this degradation maybe differ based on the complexity of the dataset, e.g., the number of classes.

\textbf{Quality of importance weights:}  In reality, empirically estimated density ratios are noisy and may deviate from the ground truth ratios. In Figure. \ref{fig:syn}(c) we simulate this setting by increasing the average amount of normal noise added to the importance weights (rest of the parameters are constant) and observing its effect on the calibration performance. We notice that with increasing noise, weighted calibrations performance performance can degrade to below uncalibrated. The sensitivity of the calibrator to noise can change with extent of domain shift.

 \subsection{EXPERIMENTS ON REAL WORLD DATA}
\label{real-data}

In this section, we evaluate calibration performance on real world datasets using classifiers trained only on source (using pre-trained ResNet50) and a range of domain adapted classifiers (LeNet-5 and ResNet50 trained using CDAN\footnote{We train using publicly available codes given by authors}, pre-trained ResNet-50). The accuracy of all the models is reported in the supplementary. We divide both source and target data into 70/30 splits (we directly use standard train/test when available). For the source, the larger split is used for training and the smaller split is used as validation for the post-hoc calibration method. For the target, the larger split is used for testing  and the smaller split is used to train the target-calibrated model. To obtain importance weights for our method we train a discriminator (2-hidden layer MLP) on domain invariant features as discussed in Section. \ref{ddre}. We perform normalization on the obtained weights, and leave experimentation with flattening and clipping for future work. We experiment with different discriminator and calibrator initializations keeping the classifier and dataset fixed and report the mean performance for 5 iterations (standard deviation in supplementary). 

%{\color{red} add 3-4 line description to talk about the source only classifier vs domain adapted}
%Photos from W and D contain the same objects, they are visually similar
%and present small domain gap. Images from A are usually dissimilar with images from W and D.

\textbf{Classifiers trained only on source:} We use the \textit{Office-31 dataset} \citep{saenko2010adapting} which is concerned with the task of object recognition. This dataset has images from four domains: Amazon images (A), Webcam (W) (low-resolution) and DSLR (high-resolution) (D), with 4,652 images and 31 categories.  We evaluate on six source to target transfer tasks \textbf{A} $\rightarrow$ \textbf{W}, \textbf{A} $\rightarrow$ \textbf{D}, \textbf{D} $\rightarrow$ \textbf{A}, \textbf{D} $\rightarrow$ \textbf{W}, \textbf{W} $\rightarrow$ \textbf{A} and \textbf{W} $\rightarrow$ \textbf{D}.  We use Imagenet pre-trained ResNet-50 as our initial classifier with the final layer replaced to output 31 classes. We re-train it on the labeled source and test it on the target. The domain-invariant feature representation for the discriminator is obtained form the final layer of the pre-trained Resnet-50 model.

\textbf{Domain adapted classifiers:} We use  Conditional Domain Adversarial Network (CDAN) \citep{long2018conditional}, a recent domain adaptation technique to train two different classifiers on different datasets mentioned here - (1) \textit{Digits dataset} consists of images form MNIST (M) and USPS (U) \citep{ganin2016domain} comprising of 10 classes, here we apply CDAN on a LeNet-5 classifier. We evaluate two source to target transfer tasks \textbf{M} $\rightarrow$ \textbf{U} and \textbf{U} $\rightarrow$ \textbf{M}.  (2) \textit{Office-Home dataset} \citep{venkateswara2017Deep} consists of images form Art (\textbf{Ar}, 2427),  Clipart (\textbf{Cl}, 4365), Product (\textbf{Pr}, 4439) and Realworld (\textbf{Rw}, 4357) (size in parenthesis) comprising of 65 classes. Here we apply CDAN on Resnet50 classifier. We evaluate 12 source to target transfer tasks shown in Table. \ref{tab:office_home}, exploring all the permutations of the four datasets. In both the datasets, we use the features obtained from the domain adapted layer of CDAN to train our discriminator.

\textbf{Discussion:} In  Table. \ref{bigtable}(a),  \ref{bigtable}(b) and  \ref{tab:office_home}.   we compare the ECE scores of our weighted calibration methods with the baselines. \an{The models here span a range of accuracy's from 30\% to 97\% on the target data and still remain uncalibrated. This shows that accounting for accuracy alone does not gurantee calibration.}   We use \textbf{bold font} to highlight results where weighted calibration outperforms the uncalibrated ECE.  In \textit{italics} we highlight weighted calibration which reduces the bias in unweighted calibration but still performs worse than uncalibrated ECE. This is in agreement with our discussion in Section. \ref{ddre}, where we note that we can only reduce the bias in using the source data for calibration but not completely eliminate. From these experiments, we observe that our proposed method helps in increasing the calibration performance considerably in number of cases such as D $\rightarrow$ A in Office-31 dataset where the ECE performance improves from 14.7\% to 6\% , M $\rightarrow$ U in MNIST-USPS dataset  where the ECE performance improves from 20.6\% to 7.7\% and Cl $\rightarrow$ Pr in Office-home dataset  where the ECE performance improves from 13.3\% to 6.4\%.

To explain the poor performance of weighted calibration on the remaining datasets, we refer to the analysis in Figure.\ref{fig:syn}. For example, in experiments involving office-31 datasets consider $A \rightarrow D$ or $A \rightarrow W$ where $A$ has considerably larger data compared to $D$ or $W$. This could have affected the importance of weight estimation (leading to overfitting of the discriminator and hence resulting in poor importance weights) or the low samples used in validation data could have itself affected both the weighted and using target labels calibration performance. In general, performance of calibration can be affected by multiple factors.

\section{CONCLUSION}

In this work, we identified that neural models, including domain adapted models, are miscalibrated under covariate shift. This indicates that existing domain adaption techniques focus on accuracy and not calibration. Existing calibration techniques fail to calibrate them or even sometimes worsen the calibration performance. This is a result of the inherent assumptions made by these techniques which fail to hold true when domain shift occurs. We propose a new method that overcomes the limitations of the existing techniques and adapts any calibration technique to work under domain shift using importance sampling. We show that with ground truth density ratios our method significantly improves the calibrator. We further implement the proposed method on real world datasets by employing a binary classifier to estimate the density ratios.  We demonstrate performance improvements on different datasets and analyze the effects of the different parameters involved. We also note that the efficacy of our method on real world datasets is limited by accuracy of the density ratio estimation process. Therefore, we observe that improving density ratio estimation is a crucial future direction of research which will help in improving calibration performance. 

%We identified that neural models, including domain adapted models are miscalibrated under covariate shift. This indicates that existing domain adaption techniques focus on accuracy and not calibration. Existing calibration techniques fail to calibrate them or even sometimes worsen the calibration performance. This is a result of the inherent assumptions made by these technique which fail to hold true when domain shift occurs. We propose a new method that overcomes the limitations of the existing techniques and adapts any calibration technique to work under domain shift. We realize a feasible framework to implement the proposed method on real world datasets.

%Proposed a method. Saw in reality difficult to get good density ratios limiting our method, and hence can only think of reducing the bias but not completely cover the gap to reach the upper bound. 
%Future work: to remain calibrated in both domains

%\newpage

%\subsubsection*{Acknowledgements}

%Use unnumbered third level headings for the acknowledgements title.
%All acknowledgements go at the end of the paper.

%\subsubsection*{References}
\bibliography{main}
\bibliographystyle{icmlfiles/icml2019}
%\newpage
%\input{supplementary}

\end{document}